\documentclass[letterpaper, 10 pt, conference]{ieeeconf}  
\usepackage{lipsum}

\usepackage{stfloats}
\usepackage{amsmath}
\usepackage{mathtools}
\usepackage{algpseudocode}
\usepackage{algorithm}
\usepackage{amssymb}
\usepackage{float}
\usepackage{multirow}
\usepackage{comment}
\usepackage{cite}
\usepackage{array}
\usepackage[utf8]{inputenc}
\usepackage[T1]{fontenc}
\usepackage{url}
\usepackage{subcaption}
\usepackage{array}
\usepackage[flushleft]{threeparttable} 
\newcolumntype{M}[1]{>{\centering\arraybackslash}m{#1}}
\usepackage{tabularx,ragged2e}

\IEEEoverridecommandlockouts                              

\title{\LARGE \bf
SomnNET: An SpO2 Based Deep Learning Network for Sleep Apnea Detection in Smartwatches
}

\author{Arlene John$^{1}$, Koushik Kumar Nundy$^{2}$, Barry Cardiff$^{3}$ and Deepu John$^{3}$
\thanks{$^{1}$Arlene John is with University College Dublin, Ireland,
        Email: {\small arlene.john@ucdconnect.ie}}%
\thanks{$^{2}$Koushik Kumar Nundy is with Think Biosolution,
        Email: {\small kknundy@thinkbiosolution.com}}%
\thanks{$^{3}$Barry Cardiff and Deepu John are with University College Dublin, Ireland, Email:
        {\small \{barry.cardiff, deepu.john\}@ucd.ie}}%
\thanks{This work was supported in part by the Irish Research Council under the New Foundations Scheme; and in part by the Microelectronic Circuits Centre Ireland under Grant MCCI-2018-03.}}

\begin{document}

\maketitle
\thispagestyle{empty}
\pagestyle{empty}

\begin{abstract}

The abnormal pause or rate reduction in breathing is known as the sleep-apnea hypopnea syndrome and affects the quality of sleep of an individual. A novel method for the detection of sleep apnea events (pause in breathing) from peripheral oxygen saturation (SpO2) signals obtained from wearable devices is discussed in this paper. The paper details an apnea detection algorithm of a very high resolution on a per-second basis for which a 1-dimensional convolutional neural network- which we termed SomnNET- is developed. This network exhibits an accuracy of 97.08\% and outperforms several lower resolution state-of-the-art apnea detection methods. The feasibility of model pruning and binarization to reduce the computational complexity is explored. The pruned network with 80\% sparsity exhibited an accuracy of 89.75\%, and the binarized network exhibited an accuracy of 68.22\%. The performance of the proposed networks is compared against several state-of-the-art algorithms.


\end{abstract}

\section{INTRODUCTION}

An increase in healthcare and diagnosis costs, along with an aging populace, has placed tremendous pressure on our healthcare systems.  Automatic diagnosis/ detection of diseases through artificial intelligence (AI) techniques in wearable devices are seen as a possible solution to tackle this issue \cite{Malasinghe, NUS_sensor}. In many cases, these AI algorithms are integrated into the wearable device itself to reduce channel bandwidth usage \cite{ICECS_Approximate,OJCAS_binary}. The abnormal reduction or pause in breathing during sleeping, associated with a reduction in blood oxygen levels is known as the sleep apnea-hypopnea syndrome \cite{Cen1}. A complete pause in breathing is termed  as apnea, and a temporary reduction in respiration rate indicated by a drop in oxygen saturation for a minimum of 10 seconds is termed as hypopnea \cite{Xie}. The diagnosis of sleep-related disorders is traditionally done through overnight polysomnography under the supervision of a clinician. Different sensors are attached to the patient's body to obtain signals for later analysis by sleep experts during polysomnography for a final diagnosis \cite{Xie}. Recording polysomnograms are costly and are not comfortable for patients. Therefore, the development of an automatic sleep-apnea detection method that is easily available, non-intrusive, and readily accessible is of paramount importance.
\par Peripheral oxygen saturation (SpO2) is an estimation of the oxygen saturation level in the blood usually measured with a pulse oximeter device, which is non-invasive and is usually found in smartwatches. Methods for analyzing obstructive sleep apnea events from pulse oximeter data were reviewed in \cite{Terrill}. Various methods for sleep apnea detection based on deep learning were reviewed in \cite{Mostafa}. In literature, most sleep-apnea detection algorithms using deep learning methods exhibit a resolution of 1 minute ie., inferences are obtained on a per-minute basis \cite{Mostafa2, Almazaydeh, Xie}. Moreover, SpO2 based sleep apnea detection algorithms are often accompanied by other signals to improve inferences \cite{Xie, Cen1}. The highest resolution of sleep apnea detection from SpO2 signals (in combination with other signals) was explored in \cite{Cen1} with a resolution of 1 second and with a performance accuracy of 79.61\%, which leads to the novelty of this work: \\
1. Development of a per-second sleep apnea detection algorithm from single-lead peripheral oxygen saturation data which is suitable for deployment in smartwatches and with a resolution higher than most state-of-the-art methods.\\ 
2. Development of a 1-dimensional convolutional neural network (1D-CNN) for this task, which we termed SomnNET (for \textit{somnum} net), which reduces the need for feature extraction stages and the task of identifying features useful for apnea detection. \\
3. Complexity analysis of the developed network and complexity optimization using network pruning methods and binarization\footnote{ Code available at https://github.com/arlenejohn/CNN\_Sleep\_apnea\_SpO2.}.\\

\section{Methodology}
\subsection{Method Outline}
A method for sleep apnea detection on a per-second basis is proposed in this article. The data preparation for per-second apnea detection using peripheral oxygen saturation signals is carried out as discussed in \cite{Arlene_iscas2021}. A sample of the training data contains a single signal window of 11 seconds with the 2\textsuperscript{nd} second corresponding to the event being detected (an apnea or non-apnea event) as discussed in \cite{Arlene_iscas2021}. Windows that contained SpO2 values less than 50\% were considered artifacts and were dropped from the dataset. 
\begin{figure}[h]
  \centering
  \includegraphics[width=0.49\textwidth,keepaspectratio]{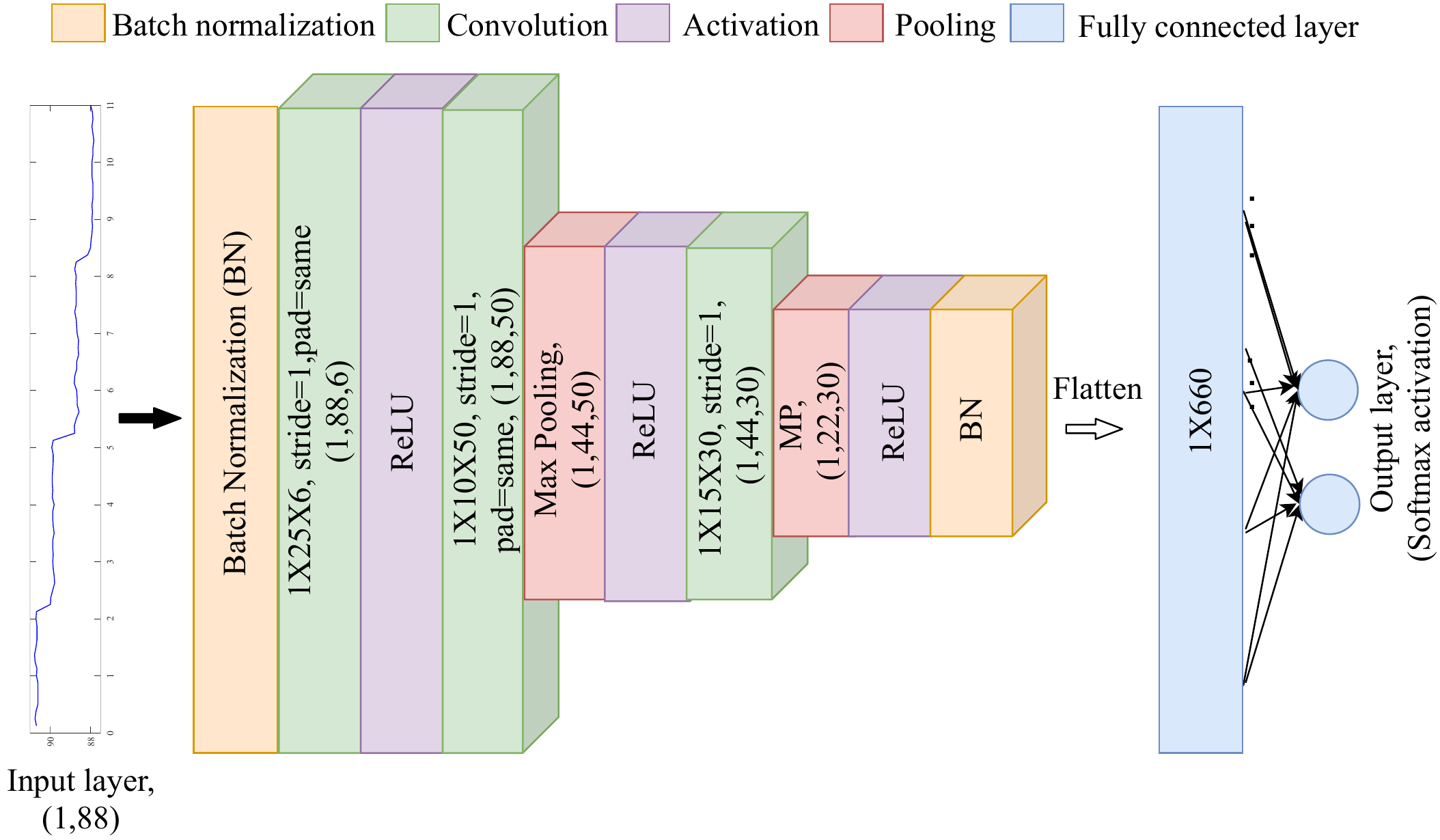}
  \caption{The architecture of SomnNET which we propose for per-second sleep apnea detection.}
  \label{fig1}
\end{figure}
\subsection{Dataset}
\begin{figure}[h]
  \centering
  \begin{subfigure}{.2\textwidth}
  \centering
  \includegraphics[width=\textwidth,keepaspectratio]{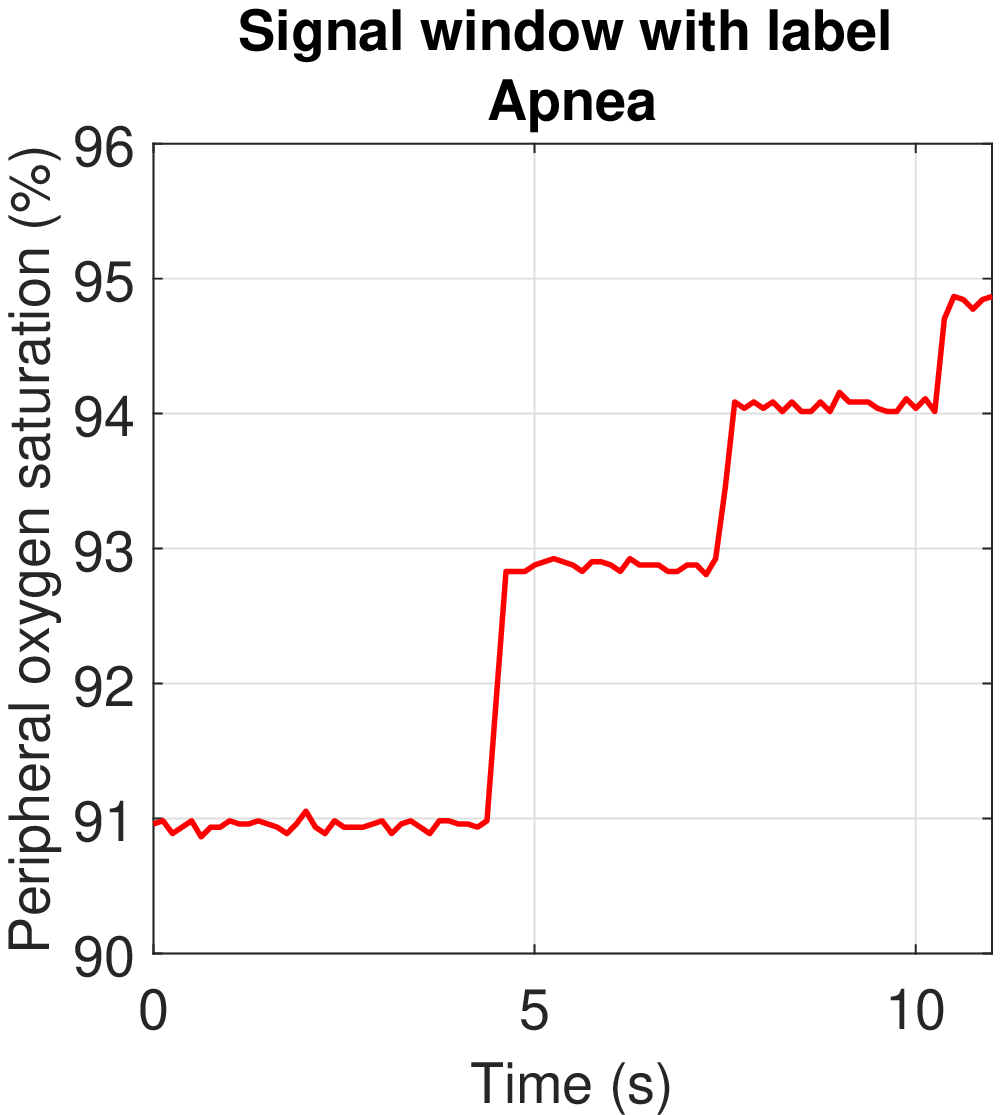}
\end{subfigure}%
\hspace{0.015\textwidth}
\begin{subfigure}{.2\textwidth}
  \centering
  \includegraphics[width=\textwidth,keepaspectratio]{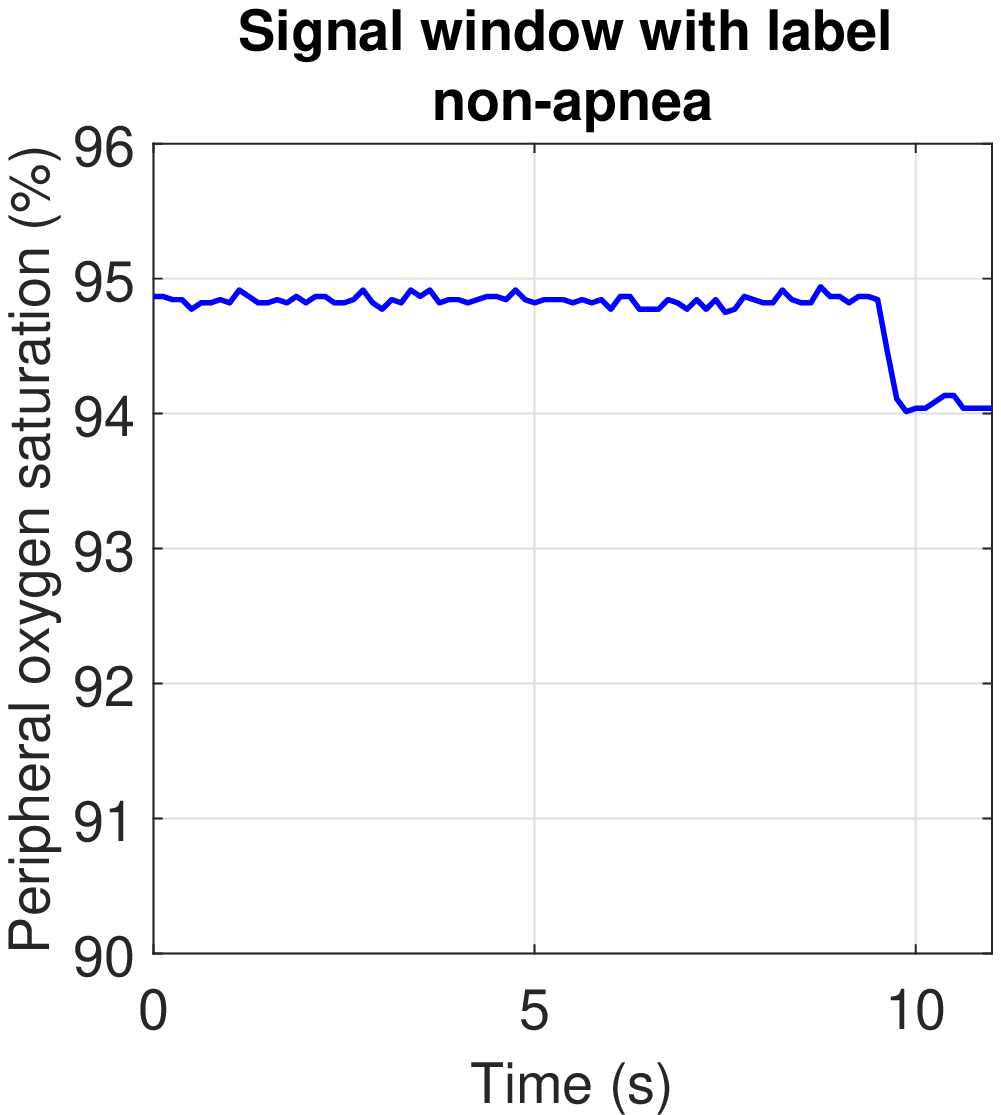}
\end{subfigure}
  \caption{(Left) SpO2 signal window with label apneic and (Right) SpO2 signal window with label non-apneic.}
  \label{fig2}
\end{figure}

In this article, the UCD St. Vincent's University Hospital's sleep apnea database containing polysomnogram records from 25 patients with a duration of 6-8 hours and with annotations for every second is used \cite{vincent}. We use the peripheral oxygen saturation (SpO2) signals which were acquired at a sampling rate of 8 Hz for this network. Fig. \ref{fig2} shows an example of signal windows where the patient is apneic and is non-apneic (record ucddb002), and it can be observed how the SpO2 levels are lower in the window depicting the apneic event. Training, validation, and test set splits are carried out epoch-wise, in the ratio of 8:1:1. Oversampling of the minority apneic class was carried out to balance the training and validation sets. Patient records without any apnea events (ucddb008, ucddb011, ucddb013, and ucddb018) were discarded.

\subsection{1D-CNN}
\label{generic}
SomnNET has 88 (8 Hz sampling frequency) input nodes, and a batch-normalization stage follows the input layer. The CNN architecture uses three convolution layers with filter lengths 25, 10, and 15, with 6, 50, and 30 of them in each layer respectively. The strides of all the convolutional layers were maintained as 1, and zero padding was used. The convolutional neural network architecture is as described in Fig. \ref{fig1}. The network parameters and training parameters are:
\begin{enumerate}
    \item Dropout in fully connected layer- probability $=0.25$,
    \item Activation function in output layer- Softmax,
    \item Activation functions- ReLU,
    \item Optimizer- ADAM,
    \item Loss function- Binary cross-entropy, and 
    \item Regularizer- L2 regularizer in output layer.
\end{enumerate}
The network generated has a total of 27,182 parameters, and for consistency throughout the paper, we refer to this network as Model 1. 

\subsection{Pruning}
 The systematic removal of parameters from an existing network is a popular approach for reducing resource requirements at prediction time and is referred to as pruning\cite{blalock}. Here, magnitude-based weight pruning is used, which gradually zero out model weights during the training process. This enables compression of the model and is suitable for deployment in resource-constrained devices like smartwatches. We use SomnNET and attempt to sparsify the network with sparsity varying from 10\% to 80\%, and the pruned network is referred to as Model 2.

\subsection{Binarized CNN}
Courbariaux \textit{et al.} proposed an algorithm to develop neural networks with binary weights and activations\cite{Matthieu}. Binarized kernel elements and weights substantially improve power efficiency by reducing memory size and accesses. In this paper, the kernel parameters in the convolution layers and the weights in the fully connected layers of SomnNET are binarized. Since the activation functions used are ReLU, we do not attempt to binarize the activation functions. This network has 27,094 parameters as the bias terms in Model 1 are eliminated. The binarized SomnNET is referred to as Model 3.


\section{Results}
\subsection{Performance Analysis}
A method of training over the full dataset and simultaneous validation on the validation set for each epoch was used for generating SomnNET. A validation callback was carried out to track the set of weights that exhibited the highest validation accuracy during training, and these weights were chosen as the final network weights. Model 1 exhibited an accuracy of 97.08\%, a specificity of 97.42\%, and a sensitivity of 84.65\% on the test set. The performance is found to outperform other state-of-the-art methods, which could be attributed to a suitable 1D CNN network that was developed along with the regularization methods used, and the validation callback ensuring that the final weights are not based on the training data, thereby preventing overfitting. The performance parameters are detailed in Table \ref{table2}.

\par
\begin{figure}[h]
  \centering
  \includegraphics[width=0.37\textwidth,keepaspectratio]{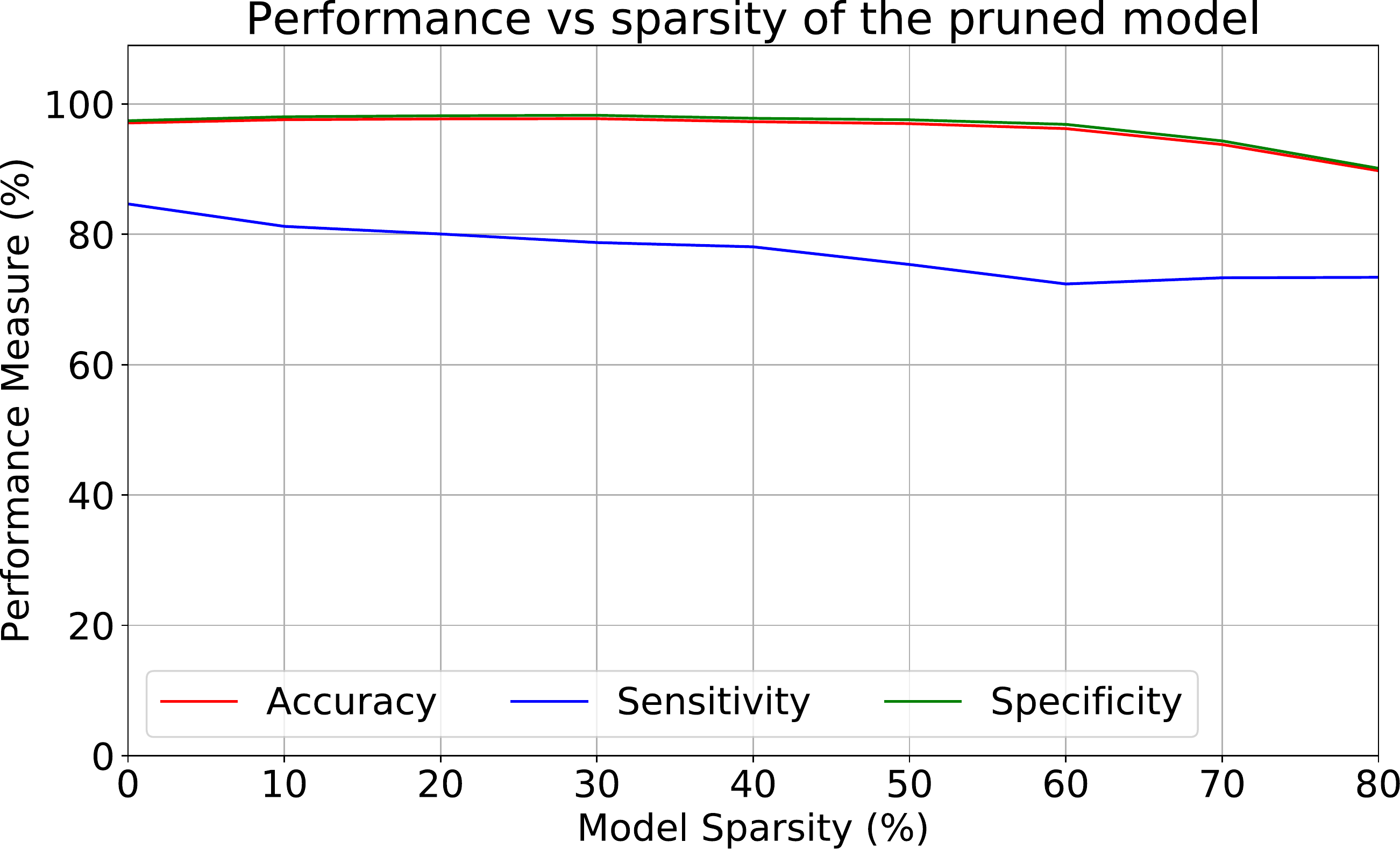}
  \caption{The performance parameters (Accuracy, Sensitivity, and Specificity) of the pruned network on the test set when the sparsity of Model 2 is increased from 10\% to 80\%.}
  \label{fig4}
\end{figure}

 The performance of the pruned networks when the sparsity of SomnNET is at 10\% to 80\% is observed. In the pruned networks, the complexity is reduced by increasing the sparsity to the desired levels in the convolutional layers and the fully connected layer. The performance parameters of the pruned networks for different sparsity levels are shown in Fig. \ref{fig4}. The performance drops with an increase in sparsity levels as is expected, and therefore an optimal sparsity level can be achieved by deciding a trade-off between power/resource consumption and performance. However, it can be observed that accuracy and specificity increase when the network is sparsified by 10\%, but it is accompanied by a significant drop in sensitivity. Even though accuracy increases, the weights that make the network sensitive are pruned away when sparsity is increased to 10\% here. We also observe that sensitivity increases when sparsity is increased to 70\% and 80\%. It may appear that even though overall performance drops with a drop in accuracy, the weights that make the network highly specific are pruned away with an increase in sparsity, making the network more sensitive. The pruned SomnNET exhibited an accuracy of 89.75\%, a specificity of 90.19\%, and a sensitivity of 73.39\% on the test set consisting of data from all the patients at 80\% network sparsity. 
\par The binarized SomnNET model was trained in a similar manner as Model 1. This network exhibited an accuracy of 68.22\%, a specificity of 67.94\%, and a sensitivity of 78.44\% on the test set. The performance of the binarized network is poor when compared to Model 1 and Model 2. When compared to Model 2 (at 80\%) in terms of sensitivity alone, Model 3 performs better at the cost of accuracy. An investigation into various combinations of binarized layers and non-binarized layers is required to achieve the desired performance levels at low computational complexity.

\begin{figure}[h]
  \centering
  \includegraphics[width=0.37\textwidth,keepaspectratio]{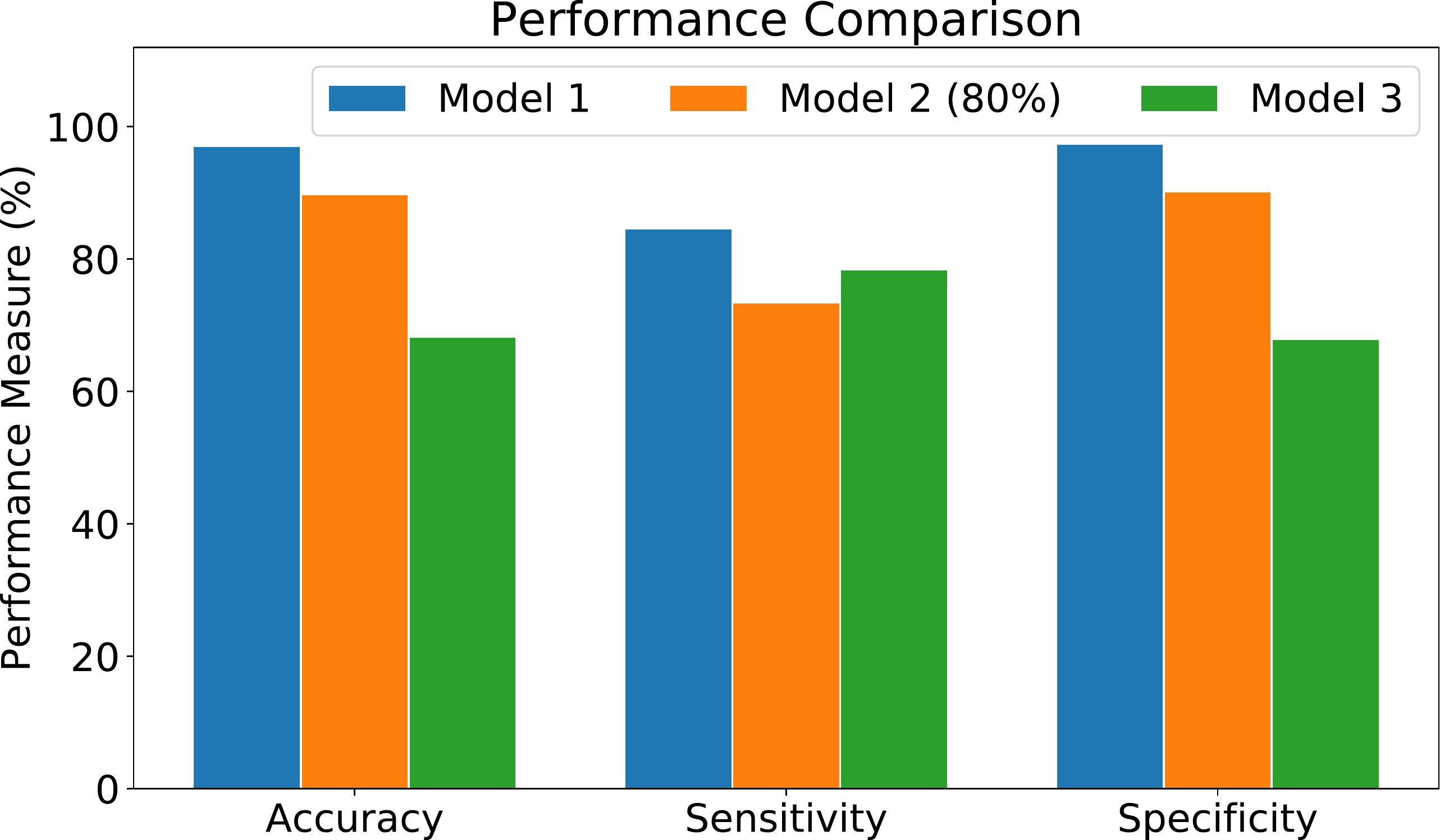}
  \caption{Bar plots of Accuracy, Specificity,  and Sensitivity of the three networks Model 1, Model 2, and Model 3 for comparison on the test set.}
  \label{fig3}
\end{figure}

\begin{table}[]
\scriptsize
\caption{Performance of the three networks Model 1, Model 2, and Model 3 in terms of accuracy, sensitivity, and specificity}
\centering
\begin{tabular}{|c|c|c|c|}
\hline
\textbf{Model} & \textbf{Accuracy (\%)} & \textbf{Sensitivity (\%)} & \textbf{Specificity (\%)} \\ \hline
Model 1    & 97.08         & 84.65            & 97.42           \\
\hline
Model 2 (10\%)    & 97.58         & 81.21            & 98.02             \\ \hline
Model 2 (20\%)    & 97.70         & 80.04            & 98.18             \\ \hline
Model 2 (30\%)    & 97.73         & 78.73            & 98.25             \\ \hline
Model 2 (40\%)    & 97.27         & 78.07            & 97.79             \\ \hline
Model 2 (50\%)    & 96.97         & 75.36            & 97.56             \\ \hline
Model 2 (60\%)    & 96.21         & 72.37            & 96.86             \\ \hline
Model 2 (70\%)    & 93.77         & 73.32            & 94.32             \\
\hline
Model 2 (80\%)    & 89.75         & 73.39            & 90.19             \\
\hline

Model 3    & 68.22         & 78.44            & 67.94  \\
\hline

\end{tabular}

\label{table2}
\end{table}
\begin{table}[]
\scriptsize

\centering
\caption{Computational Complexity in terms of multiplication and addition operations and the energy consumption of the networks during prediction}
\begin{tabular}{|c|c|c|c|c|}
\hline
\textbf{Network} & \textbf{Params} & \textbf{Mul} & \textbf{Add} & \textbf{Energy ($\mu$J)} \\  \hline
Model 1    & 27182      & 1270016          & 1272876    & 0.4964      \\ \hline
Model 2 (10\%)   &  24485      & 1143428          & 1146288    & 0.4470      \\ 
 \hline
Model 2 (20\%)   &  21788      & 818840          & 821700    & 0.3205      \\ 
 \hline
Model 2 (30\%)   &  19091      & 889724          & 892584    & 0.3481      \\ 
 \hline
Model 2 (40\%)   &  16394      & 762608          & 765468    & 0.2985      \\ 
 \hline
Model 2 (50\%)   &  13697      & 636020          & 638880    & 0.2492      \\ 
 \hline
Model 2 (60\%)   &  11000      & 508904          & 511764    & 0.1996     \\  \hline
Model 2 (70\%)   &  8303     & 382316         & 385176    & 0.1502     \\
 \hline
Model 2 (80\%)   &  10106      & 255200         & 258060    & 0.1006      \\  \hline

Model 3    & 27094      & 1496          & 1179946    & 0.0236  \\ \hline

\end{tabular}
\label{table3}
\end{table}

\par The performance of the various versions of SomnNET (Model 2 at 80\%) in terms of accuracy, sensitivity, and specificity are compared in Fig. \ref{fig3}. The performance parameters of all the networks are detailed in Table \ref{table2}. From the table, it can be observed that Model 1 and Model 2 are suitable for per second apnea detection in smartwatches, while Model 3 requires further investigation. 

\begin{table*}[ht]
\scriptsize
\centering
\caption{Comparison of state-of-the-art SpO2 record based sleep apnea detection algorithms with the performance of the proposed 1D-CNN based models for apnea detection.}
\centering
\begin{tabular}{|c|c|c|c|c|c|c|c|}
\hline
\textbf{Article}                 & Mostafa \textit{et al.} \cite{Mostafa2}                                                                                                 & Almazayadeh \textit{et al.} \cite{Almazaydeh}                                                                & Xie \textit{et al.} \cite{Xie}                                                                         & Cen \textit{et al.} \cite{Cen1}                                                                                            & \multicolumn{3}{c|}{This work}                                                            \\ \hline
\textbf{Dataset}                 & \begin{tabular}[c]{@{}c@{}}Physionet Apnea ECG\\ database and UCD St. Vincent's\\ database\end{tabular} & \begin{tabular}[c]{@{}c@{}}Physionet Apnea ECG\\ database\end{tabular} & \begin{tabular}[c]{@{}c@{}}UCD St. Vincent's\\ database\end{tabular}            & \begin{tabular}[c]{@{}c@{}}UCD St. Vincent's\\ database\end{tabular}                               & \multicolumn{3}{c|}{\begin{tabular}[c]{@{}c@{}}UCD St. Vincent's\\ database\end{tabular}} \\ \hline
\textbf{Signal}                  & SpO2                                                                                                    & SpO2                                                                   & \begin{tabular}[c]{@{}c@{}}SpO2 and \\ electrocardiogram\end{tabular}           & \begin{tabular}[c]{@{}c@{}}SpO2, oronasal airflow,\\ ribcage and abdominal\\ movement\end{tabular} & \multicolumn{3}{c|}{SpO2}                                                                 \\ \hline
\textbf{Resolution}              & 1 minute                                                                                                & 1 minute                                                               & 1 minute                                                                        & 1 second                                                                                           & \multicolumn{3}{c|}{1 second}                                                             \\ \hline
\multirow{2}{*}{\textbf{Method}} & \multirow{2}{*}{3 layer deep belief network}                                                            & \multirow{2}{*}{Multilayer neural network}                             & \multirow{2}{*}{\begin{tabular}[c]{@{}c@{}}Bagging with\\ RepTree\end{tabular}} & \multirow{2}{*}{2D CNN}                                                                            & \multicolumn{3}{c|}{SomnNET}                                                               \\ 
                        &                                                                                                         &                                                                        &                                                                                 &                                                                                                    & Model 1     & \begin{tabular}[c]{@{}c@{}}Model 2 \\ (80\%)\end{tabular}    & Model 3     \\ \hline
\textbf{Accuracy}                & 97.64 \%                                                                                                & 93.30 \%                                                               & 84.40 \%                                                                        & 79.61\%                                                                                            & 97.08 \%    & 89.75 \%                                                     & 68.22 \%    \\ \hline
\textbf{Sensitivity}             & 78.75 \%                                                                                                & 87.50 \%                                                               & 79.75 \%                                                                        & -                                                                                                  & 84.65 \%    & 73.39 \%                                                     & 78.44 \%   \\ \hline
\end{tabular}
\label{table4}
\end{table*}
\subsection{Computational Complexity Analysis}
\par The computational complexities of the three networks were calculated in terms of the number of multiplications and additions required for each second \cite{Arlene_iscas2021}. Detection of sleep apnea events with SomnNET (Model 1) requires 1270016 multiplications and 1272876 additions. In the case of Model 2, the gains due to pruning can be approximated by estimating the number of operations with non-zero numbers. When SomnNET is at 80\% sparsity, the number of computations required involves 255200 multiplications and 258060 additions respectively. Detection of apnea events with the binarized SomnNET network requires just addition operations at the convolution layers and fully connected layer, since the weights are either +1 or -1, and therefore Model 3 requires 1496 multiplications and 1272876 additions. The total energy consumption during prediction is found to be 0.4964 $\mu$J, 0.1006 $\mu$J, and 0.0236 $\mu$J for Model 1, Model 2, and Model 3 respectively. This is estimated by assuming that the energy required for a 16-bit multiplication accumulation (MAC) operation is 0.39 pJ \cite{cite10, cite12}, and for a 16-bit adder is around 20 fJ \cite{Taco2} in 28nm FD-SOI technology. The complexity in terms of multiplications and additions and the corresponding energy consumption are discussed in Table \ref{table3}.

\par The performance of the proposed 1D-CNN  networks is compared with that of state-of-the-art algorithms in Table \ref{table4}. It can be seen that Model 1 and Model 2 exhibit comparable performance in terms of accuracy but higher sensitivity to methods proposed in \cite{Mostafa2} which is tested on a combined database, and therefore a direct comparison to our method is not possible. Model 1 and Model 2 outperform the method proposed in \cite{Almazaydeh} in terms of accuracy, which is tested on the Physionet Apnea Database. Model 1 and Model 2 outperforms the methods proposed in \cite{Xie} and \cite{Cen1} in terms of accuracy and sensitivity. Direct comparison with \cite{Xie} and \cite{Cen1} is possible due to the same dataset being used. It is also noteworthy that the method proposed in this work uses a single sensor source, SpO2, while in \cite{Xie} and \cite{Cen1} a combination of different signals is used for inference, and yet our method outperforms these two works. This could be because in \cite{Xie}, bagging with repTree is used along with 1-minute resolution, which significantly reduces the number of training samples and due to the hand-engineered features. In the case of \cite{Cen1}, a 2D CNN with a combination of different signals is used. The model complexity and larger number of learnable parameters could have caused the model to overfit to the training set, leading to poor performance on the test set.\\

\section{Conclusions}
In this article, we analyze the feasibility of sleep apnea detection from SpO2 signals on a per-second basis. The requisite features for apnea event detection are learned by the proposed 1D-CNN network, which is termed as SomnNET. Two strategies to reduce the network size to make it suitable for implementation in smartwatches are analyzed, and the performance of these two methods is studied. The proposed SomnNET network achieved an accuracy of 97.08\%, the pruned SomnNET network at 80\% sparsity achieved an accuracy of 89.75\%, and an accuracy of 68.22\% was exhibited by the binarized SomnNET. These networks can work at a high resolution (per-second) for any subject with minimal tuning and is suitable for implementation on smartwatches. In future works, the filters learned by the 1D-CNN layers of SomnNET can be analyzed to explain the feature extraction process. The correlation with human understanding of requisite filter banks for SpO2 based apnea detection with the filter weights learned by the CNN can be analyzed to understand its impact on the model performance.

\bibliographystyle{IEEEtran}
\bibliography{bibliography}

\end{document}